# UniSino: Physics-Driven Foundational Model for Universal CT Sinogram Standardization

Xingyu Ai, Shaoyu Wang, Zhiyuan Jia, Ao Xu, Hongming Shan, Jianhua Ma, Qiegen Liu

***Abstract*—During raw-data acquisition in CT imaging, diverse factors can degrade the collected sinograms, with undersampling and noise leading to severe artifacts and noise in reconstructed images and compromising diagnostic accuracy. Conventional correction methods rely on manually designed algorithms or fixed empirical parameters, but these approaches often lack generalizability across heterogeneous artifact types. To address these limitations, we propose UniSino, a foundation model for universal CT sinogram standardization. Unlike existing foundational models that operate in image domain, UniSino directly standardizes data in the projection domain, which enables stronger generalization across diverse undersampling scenarios. Its training framework incorporates the physical characteristics of sinograms, enhancing generalization and enabling robust performance across multiple subtasks spanning four benchmark datasets. Experimental results demonstrate thatUniSino achieves superior reconstruction quality both single and mixed undersampling case, demonstrating exceptional robustness and generalization in sinogram enhancement for CT imaging. The code is available at: https://github.com/yqx7150/UniSino.***

## I. INTRODUCTION

Computed tomography (CT) is a cornerstone of modern diagnostic imaging [1], enabling the reconstruction of high-resolution cross-sectional images of internal anatomy through the acquisition of multi-view projection data and nonlinear physical reconstruction processes. In CT imaging, the sinogram represents the raw projection data before image reconstruction. However, in practical clinical data acquisition, raw sinogram data are frequently degraded by a multitude of factors—including hardware limitations, environmental variability, and patient-induced factors—resulting in complex and heterogeneous data corruption [2]. Without effective preprocessing, such imperfections are readily amplified through the reconstruction process, which lead to severe image artifacts [3], including detector-induced ring patterns [4], beam

X. Ai, S Wang, Q. Liu and Z. Jia are with School of Information Engineering, Nanchang University, Nanchang, China ({aixingyu.aiden, wangshaoyu, liuqiegen}@ncu.edu.cn; jiazhiyuan@email.ncu.edu.cn)

A. Xu is with School of Mathematics and Computer Sciences, Nanchang University, Nanchang, China (xuao@email.ncu.edu.cn)

H. Shan is with Institute of Science and Technology for Brain-inspired Intelligence, MOE Frontiers Center for Brain Science, and Key Laboratory of Computational Neuroscience and Brain-Inspired Intelligence, Fudan University, Shanghai 200433, China.( hmshan@fudan.edu.cn)

J. Ma is with Key Laboratory of Biomedical Information Engineering of Ministry of Education, School of Life Science and Technology, Xi'an Jiaotong University, Xi'an, Shaanxi, China. (jhma@smu.edu.cn)

hardening from metal implants [5], geometric distortions from miscalibration [6], and motion-induced inconsistencies [7, 8]. These artifacts not only degrade visual quality but critically compromise diagnostic reliability. To meet growing demands for rapid imaging and dose reduction, contemporary CT systems increasingly rely on acquisition protocols such as limited-angle (LA) [9], sparse-view (SV) [10], and low-dose (LD) scanning [11]. While these strategies successfully mitigate radiation exposure and improve throughput [12-14], they further exacerbate data incompleteness and noise, giving rise to streak artifacts, structural collapse, and high-frequency degradation. The coexistence and interplay of these diverse artifacts pose a formidable challenge to conventional reconstruction algorithms. Notably, many of these imperfections manifest in combination, rendering handcrafted or task-specific correction techniques insufficient [15]. Despite advances in algorithmic correction, most existing CT systems still rely on proprietary, workflow-intensive pre-processing software that often requires domain-specific tuning and extensive manual parameter adjustment. Non-standardized data exacerbate this issue, as downstream processing algorithms must be carefully calibrated for specific instruments and cannot be easily generalized. These limitations underscore the urgent need for a unified, generalizable, and operationally efficient framework for CT sinogram standardization. A standardized model has the potential to convert various types of under-sampled data into clean and consistent sinograms, facilitating subsequent processing and improving overall workflow efficiency.

In recent years, deep learning has demonstrated remarkable success across various CT imaging tasks, including LA reconstruction [16], SV completion [17], LD noise suppression [18], metal artifact reduction [19], ring artifact correction [20], and motion compensation [21]. However, these models are typically tailored for single, well-defined tasks and trained on narrowly scoped datasets, limiting their generalizability to diverse clinical scenarios. Most current approaches are designed for specific defect types and imaging setups [22-24], resulting in isolated model architectures that hinder crosstask knowledge sharing and scalability. This fragmentation not only increases the computational burden and maintenance cost but also reduces deployment efficiency in clinical settings. Moreover, these task-specific models are often sensitive to changes in imaging protocols or acquisition parameters, necessitating retraining for each new configuration [25-27].

With the rapid development of general artificial intelligence, foundational models (FMs) have shown unprecedented cross-task capabilities [28]. Through large-scale pretraining on diverse datasets, these models enable effective generalization across modalities and application domains. Foundational models have already demonstrated breakthrough performance in natural language processing [29, 30], computer vision [31, 32], and vision-language tasks [33-35], facilitating multitask learning and knowledge transfer. In medical imaging, however, foundational model research remains in its early stages, with existing efforts primarily focused on downstream tasks such as image segmentation and classification [32, 36-40]. Very little work has explored foundational modeling for raw CT projection data [41], especially in the context of sinogram standardization. Several key challenges hinder progress in this direction: a) Data limitations. Ethical constraints, commercial protection, and hardware restrictions make it difficult to obtain large-scale, high-quality datasets containing both defective and corresponding standardized sinogram pairs. Data curation typically requires extensive domain expertise and significant manual effort, which severely restricts the feasibility of training generalized models. b) Modeling complexity. Raw CT sinograms are not intuitively interpretable and often lack direct anatomical representation. Fig. 1 illustrates the types of undersampling. In this study, the undersampling types are categorized according to their causes into those arising from non-standard scanning protocols and those resulting from non-ideal reconstruction conditions. The presence of a wide variety of artifact types and undersampling patterns makes it difficult to design models with strong generalization capabilities. c) Task complexity. Sinogram standardization involves a range of subtasks, including correction, enhancement, artifact suppression, and data completion. These inherently form a multi-task, multi-modal learning challenge, while most existing models are built for isolated tasks with limited representation capacity.

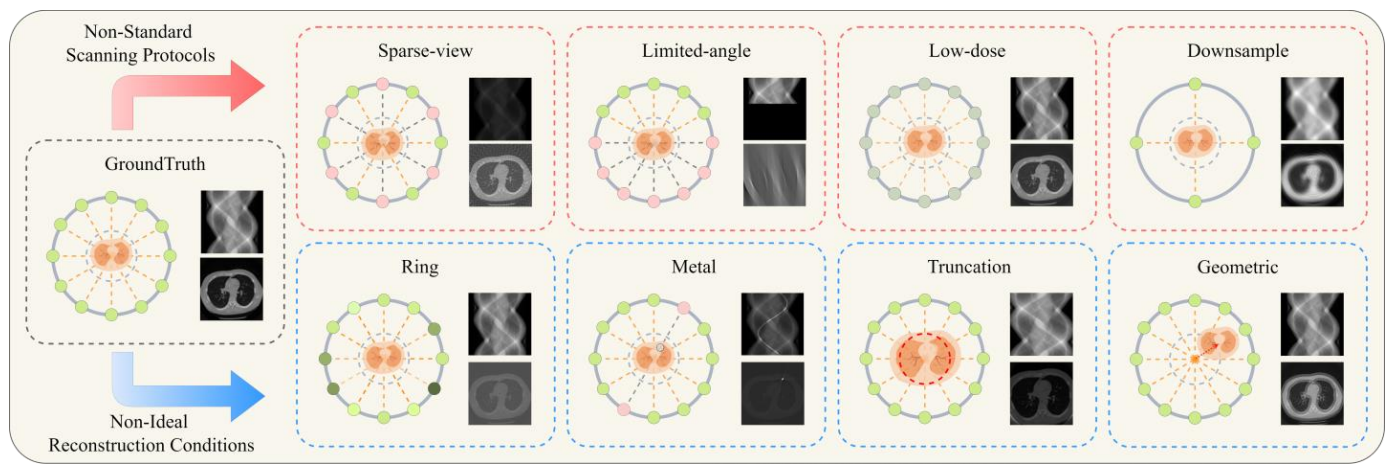

**Fig. 1.** Illustration of undersampled data. Part of the degradation arises from non-standard scanning protocols, while another part is caused by unavoidable hardware-related issues and other non-ideal scanning conditions that lead to reduced imaging quality.

In this work, we propose a general foundation model framework for CT sinogram standardization, termed the Universal CT Sinogram Standardization Foundational Model (UniSino). The core of UniSino is a foundational model designed to encode physical knowledge and adapt to diverse CT preprocessing tasks. Fig. 2 illustrates the clustering of samples in the projection and image domains.

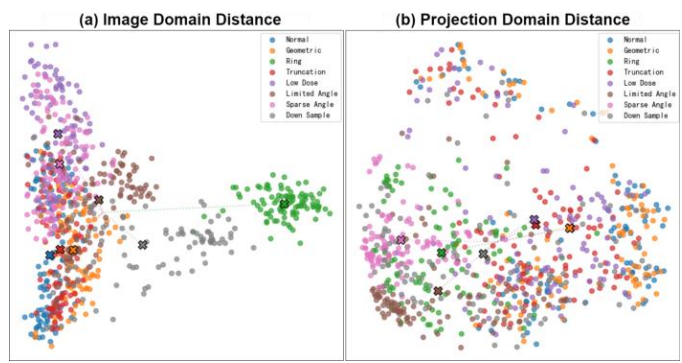

**Fig.2.** Scatter plots of data distributions across different domains encoded by InceptionV3. Figure (a) shows that samples in the image domain exhibit clustering by undersampling category, while the differences across categories are significant. Figure (b) shows that different undersampling categories in the projection domain exhibit better consistency and a more uniform distribution.

Working directly in this projection domain offers a key advantage: errors or non-standard signals corrected here can be addressed at their source, whereas if left untreated they are amplified during the nonlinear back-projection process and manifest as severe artifacts in the image domain. Moreover, image domain samples tend to cluster by artifact type, which makes it difficult for a model to learn a unified distribution across heterogeneous degradations. In contrast, non-standard sinograms exhibit a more dispersed and statistically uniform distribution, providing a smoother basis for model training and stronger cross-task generalization. These properties make the projection domain a natural choice for standardization, ensuring more reliable data quality before reconstruction and reducing the propagation of errors into the final CT images. The model leverages perceptual compression, which enables it to handle various types of non-standard sinograms with a relatively small number of parameters. Meanwhile it also employs physically-informed training objectives to efficiently learn generalized sinogram enhancements.

The main contributions of this paper can be summarized as follows:

- We introduce UniSino as the first foundational model for CT sinogram standardization. It is capable of handling a wide range of preprocessing tasks, including ring artifact, LD noise removal, geometric error correction, and LA/SV completion. With its strong generalization capacity, UniSino reduces the cost of model development and improves clinical deployment efficiency.

- UniSino incorporates perceptual compression to enhance training efficiency and a diffusion framework to improve the quality of latent space. Specifically, the compressed latent representation is divided into a full-frequency pathway, which preserves the majority of image information, and a high-frequency pathway, which emphasizes artifact-related features to provide targeted latent attention. It further integrates a physics-guided projection domain loss function, which constrains the output of the model through sinogram consistency and bounded variation. These features enabling parameter-efficient adaptation and multitask generalization.

- We conduct extensive experiments on eight CT datasets. Our results demonstrate that UniSino performs robustly across single-task and mixed-task scenarios, delivering superior reconstruction quality and excellent generalization in sinogram enhancement for CT imaging.

## II. RESULTS

### A. UniSino Overview

UniSino constitutes a foundation model for CT projection domain reconstruction. Fig. 3 presents the overall architecture of UniSino. It integrates two principal components: a sinogram variation autoencoder (SinoVAE) module that learns physically constrained latent representations and a latent refinement diffusion (LRD) module that performs conditional refinement of degraded sinograms. The SinoVAE module implements an encoder-decoder structure featuring dual decoding pathways for global and edge-specific reconstruction. Training incorporates physics-constrained reconstruction losses enforcing projection domain priors, and adversarial discrimination for representation enhancement.

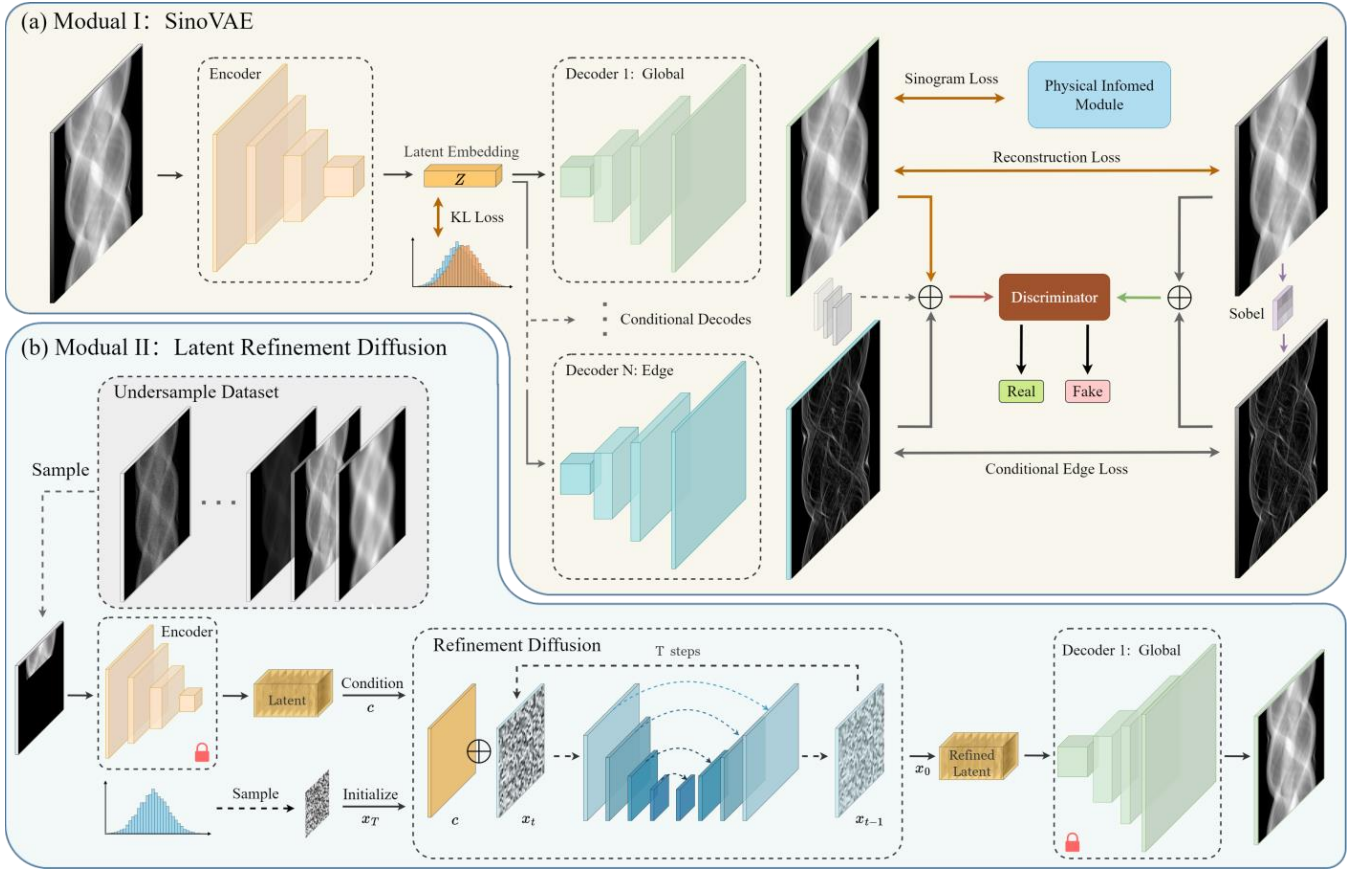

**Fig.3.** The UniSino Framework: In the first stage, SinoVAE trains an encoder–decoder pair with standardized data to achieve compression and reconstruction of input features. The training process incorporates a physics-guided module, where SinoLoss constrains the reconstruction results. In the second stage, LRD module compresses undersampled data using the pre-trained encoder, and a conditional diffusion model is employed to optimize the latent space. The decoder then reconstructs the standardized sinogram from the optimized latent representation.

The physics-guided sinogram loss (SinoLoss) is designed with reference to the imaging principles of sinograms, achieving improved consistency. The LRD module operates through a conditioned diffusion process initialized with noise samples and guided by latent encodings of undersampled data, executing iterative refinement via a global decoder to produce final reconstructions.

In the first stage of training, the SinoVAE module is optimized through a self-supervised paradigm, where training sample pairs are constructed by generating simulated undersampled counterparts from fully sampled sinograms. The architecture employs an encoding network that leverages generative KL divergence to regularize latent embeddings, coupled with a global reconstruction decoder and an edge-specific decoder featuring conditional output modulation. The training process incorporates a primary sinogram reconstruction loss, physical constraints that enforce projection domain priors, and conditional edge preservation objectives. Fig. 4 shows the detailed implementation of the physics-guided SinoLoss. A discriminator-based adversarial learning component further refines the latent representations via real-fake sample evaluation. This optimization process preserves projection domain fidelity, thereby providing preprocessed inputs for subsequent diffusion-based refinement.

In the second stage, a conditional diffusion model is trained within the latent space established by SinoVAE. To address the intrinsic randomness of conventional unconditional diffusion models during generation and ensure the stability and reliability of restoration outcomes, UniSino incorporates the latent encoding of the undersampled sinogram as a robust condition into the reverse diffusion procedure. This conditional mechanism tightly couples the restoration process with input information, substantially enhancing the fidelity of reconstructed results.

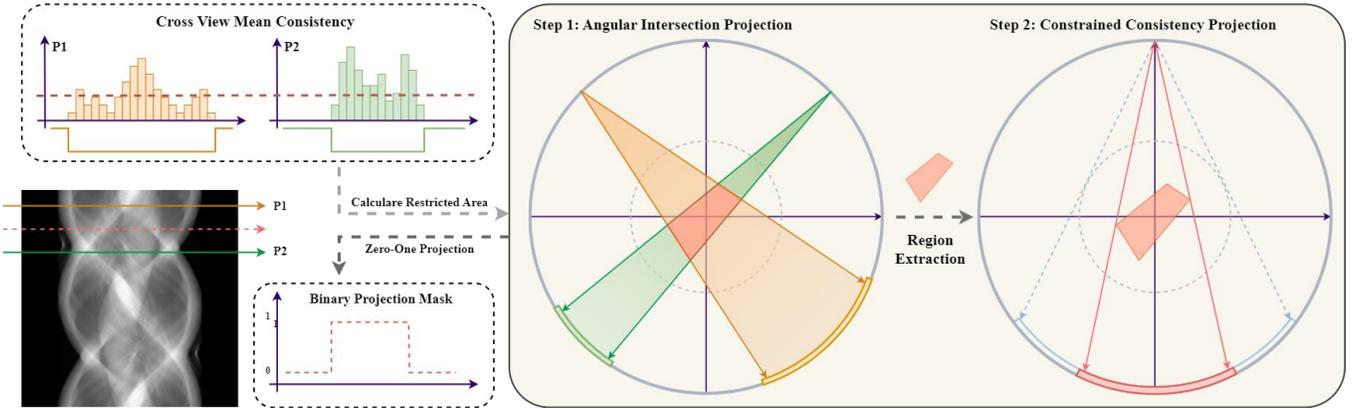

**Fig. 4.** Overview of physics-informed SinoLoss: First, since each row of data in a sinogram represents a projection of the same scanned object, the sums across different rows should remain consistent, ensuring cross-view mean consistency. Second, for a given set of observed angles, the range of possible projection values at a new angle is bounded. This boundary can be determined by performing a back-projection of the known angles to obtain an intersecting region, followed by forward-projecting along the target angle. By extracting the non-zero values from this projection, a mask is generated to identify the range where values may appear at the new angle, which is then used to constrain the projection.

The innovations of UniSino are primarily manifested in the following facets. Firstly, in the model design aspect, during the training of the SinoVAE, we integrate domain-specific physical prior knowledge to the construct a latent space that is structurally optimized for the projection domain. Unlike conventional representations, the latent embedding is explicitly decomposed into two complementary components: a full-frequency pathway, which preserves the majority of the global structural and intensity information of the sinogram, and a high-frequency pathway, which specifically encodes artifact-sensitive information such as streaks, rings, or undersampling-induced edges. This dual-pathway latent representation not only retains the essential data fidelity but also introduces targeted sensitivity to degradation patterns, allowing the model to more effectively separate true signals from artifact-related variations. The latent vectors are further refined through decoding processes with KL loss, sinogram loss, and conditional edge loss to refine the latent representations.

Secondly, in the network architecture aspect, building upon this enriched latent space from SinoVAE, we employ a dedicated LRD architecture. Unlike traditional unconditional diffusion models, LRD conditions the reverse diffusion process directly on the undersampled sinogram's latent encoding, tightly coupling the generative refinement with the actual corrupted input. This conditioning mechanism not only improves reconstruction fidelity but also enables the diffusion model to leverage the dual latent design (full-frequency + high-frequency), where the high-frequency embeddings serve as focused priors to suppress artifacts while the full-frequency embeddings ensure overall structural coherence. This tight coupling of the restoration process with input information enhances the fidelity of reconstructed results.

### B. UniSino Enables Standardization Across Multiple Types of Undersampling.

This section assesses the performance of UniSino in sinogram standardization. Experiments were conducted on the NLST lung dataset, where 300 patients were selected for undersampling, resulting in a total of 800,112 slices. Among these, 790,112 slices were used for training and 10,000 slices for testing. All slices were converted to sinograms, with each sinogram having a size of 384 × 384. The models were implemented using the PyTorch framework and optimized with the Adam optimizer. Training was performed on two NVIDIA RTX 4090 GPUs with 24 GB of memory each. The learning rate for the SinoVAE module was set to 1e-6, and for the LRD module it was set to 1e-4.

These includes detector-induced ring artifacts, metallic implant distortions, data truncation, and geometric inconsistencies. Unlike acquisition-related degradations, these artifacts are primarily linked to physical properties of the hardware, the object, or the patient.

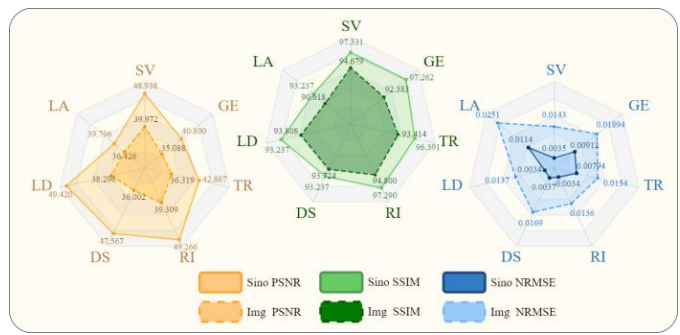

**Fig. 5.** Reconstruction metrics of UniSino in the projection-domain and image-domain under different undersampling types. The image domain results are obtained by applying FBP to the standardized sinograms. SV: sparse-view; LA: limited-angle; LD: low-dose; DS: downsample RI: ring artifact; TR: truncation artifact; GE: geometry artifact.

**Table I**
SINOGRAM STANDARDIZATION COMPARISON

| Method | PSNR | SSIM | NRMSE |
|---|---|---|---|
| CycleGAN | 27.398 | 77.046 | 0.08221 |
| ViT | 27.881 | 70.363 | 0.06632 |
| U-Net | 39.764 | 91.087 | 0.01820 |
| DDPM | 39.613 | 95.432 | 0.01385 |
| UniSino | 45.522 | 96.272 | 0.00607 |

They introduce structured corruptions into the sinogram, disrupt the expected data patterns, and present significant obstacles for reconstruction algorithms. We benchmarked UniSino against several representative methods. Table I presents the average projection-domain metrics of different methods for sinogram standardization, while Fig. 5 illustrates the performance of UniSino under different undersampling scenarios.

For projection-domain standardization, comparisons were made with U-Net and ViT. To assess its performance relative to other generative models, we further included CycleGAN and DDPM. UniSino demonstrates strong generative capability, preserving essential high-frequency details while effectively suppressing high-frequency artifacts and maintaining overall structural fidelity. In contrast, baseline methods show notable limitations. As shown in Fig. 6 and Table II, UniSino consistently outperforms competing methods across both projection and image domains. Fig. 7 further visualizes image-domain reconstructions obtained by back-projecting standardized sinograms from different models.

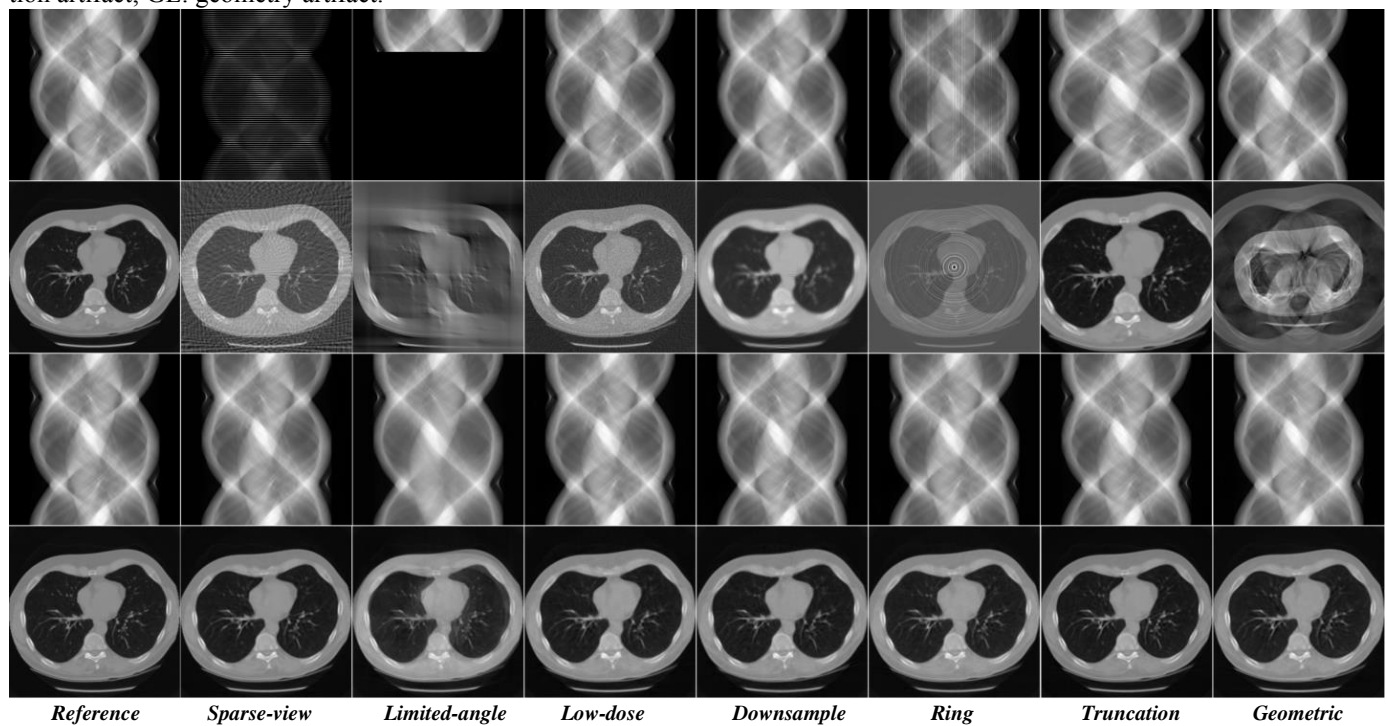

| *Reference* | *Sparse-view* | *Limited-angle* | *Low-dose* | *Downsample* | *Ring* | *Truncation* | *Geometric* |

**Fig. 6.** Visualization of UniSino standardization results. The first row shows projection-domain undersampled data, while the second row shows the corresponding image-domain undersampled data. The third row displays the reconstructed results in the projection domain, and the fourth row presents the reconstructed results in the image domain.

U-Net fails to recover fine details, yielding overly smooth outputs. DDPM achieves relatively better results, but the lack of guidance from the physical imaging mechanism leads to prominent artifacts after back-projection. Moreover, the absence of feature compression in DDPM results in substantially longer reconstruction times. Patch-based strategies reduce memory usage but often introduce block inconsistencies, causing visible artifacts and degraded metrics. In comparison, UniSino leverages an encoder-decoder architecture to compress undersampled sinograms and extract global image information, while the generative power of the diffusion model fills in missing or corrupted details. Furthermore, the inclusion of SinoLoss during decoder training reduces inconsistencies within the sinogram, thereby mitigating artifact formation.Overall, UniSino offers a unified and versatile framework for correcting diverse physical artifacts in CT sinograms. Conventional approaches typically require dedicated hardware calibration or task-specific algorithms, UniSino learns a universal transformation to a clean sinogram representation by training on a diverse dataset containing numerous instances of these condition-dependent artifacts. Its capacity to effectively remove a wide array of degradations with a single model highlights its efficiency and practical utility, offering a potent solution for harmonizing data and ensuring reliable input for image reconstruction.

**Table II**
SINOGRAM STANDARDIZATION COMPARISON OF PSNR/SSIM/NRMSE

| Method | DDPM | U-Net | CycleGAN | ViT | UniSino |
|---|---|---|---|---|---|
| Sparse-view | 43.156 / 87.110 / 0.0070 | 43.987 / 98.701 / 0.0062 | 27.973 / 91.267 / 0.0436 | 31.376 / 71.241 / 0.0270 | 48.938 / 97.331 / 0.00353 |
| Limited-angle | 21.549 / 74.661 / 0.0837 | 26.779 / 86.912 / 0.0445 | 14.533 / 54.136 / 0.1936 | 12.237 / 43.238 / 0.2444 | 39.794 / 93.237 / 0.01141 |
| Low-dose | 41.689 / 98.170 / 0.0082 | 45.736 / 98.260 / 0.0051 | 40.153 / 93.686 / 0.0098 | 35.172 / 82.611 / 0.0174 | 49.420 / 97.305 / 0.00335 |
| Down-sample | 44.385 / 91.170 / 0.0060 | 44.798 / 98.640 / 0.0057 | 40.605 / 93.593 / 0.0093 | 34.736 / 83.712 / 0.0183 | 47.567 / 95.091 / 0.00373 |
| Ring | 43.483 / 95.590 / 0.0067 | 41.242 / 97.500 / 0.0086 | 32.898 / 93.709 / 0.0235 | 35.468 / 83.591 / 0.0169 | 49.266 / 97.290 / 0.00341 |
| Truncation | 42.400 / 96.712 / 0.0076 | 37.049 / 92.311 / 0.0141 | 22.364 / 74.079 / 0.0755 | 22.821 / 69.116 / 0.0723 | 42.867 / 96.391 / 0.00794 |
| Geometric | 41.690 / 94.202 / 0.0082 | 37.705 / 95.702 / 0.0128 | 13.259 / 38.852 / 0.2202 | 23.355 / 59.036 / 0.0680 | 40.800 / 97.262 / 0.00912 |

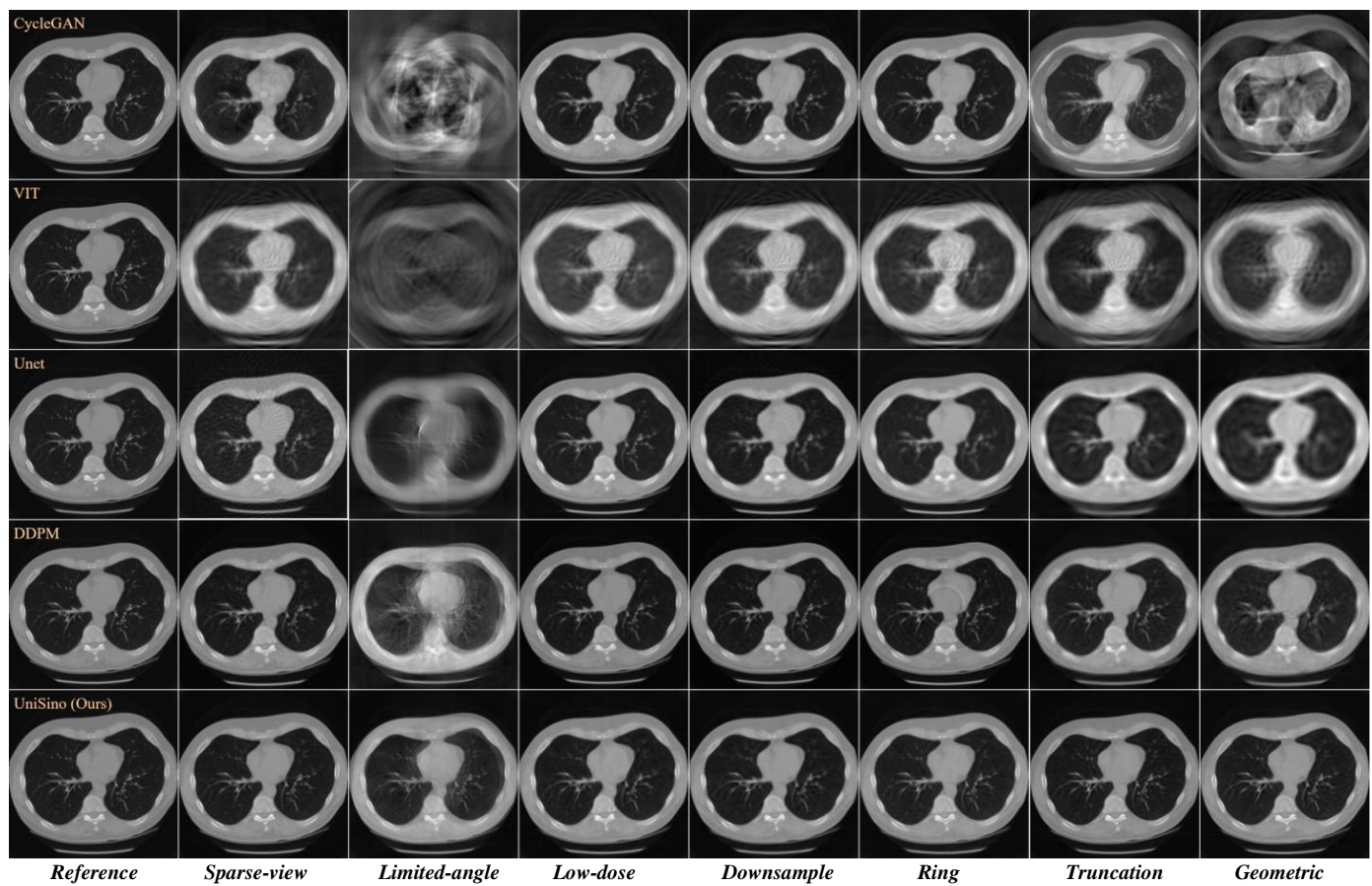

**Fig. 7.** Comparison of imaging results from different methods. Each column represents a corresponding type of undersampling. UniSino demonstrates strong consistency across various undersampling types.

### C. UniSino Enables Simultaneous Processing of Multiple Mixed Artifacts

In real clinical scenarios, sinogram degradations rarely occur in isolation, instead, they typically arise from the nonlinear interplay of multiple degradation factors. Fig. 8 shows that UniSino effectively handles diverse mixed artifacts, highlighting its robustness and clinical applicability.

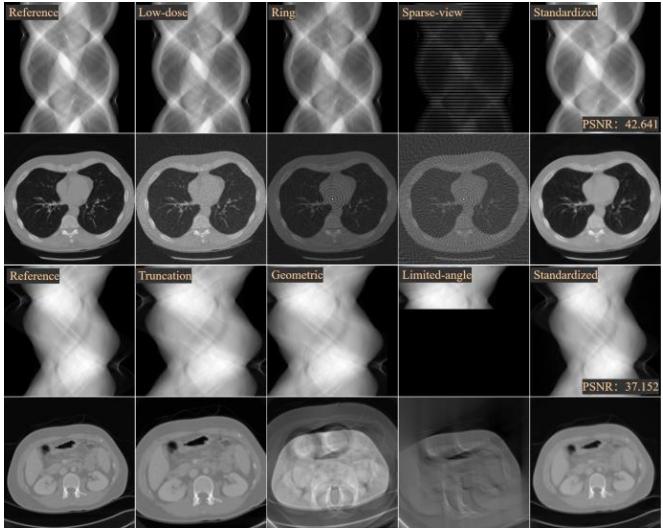

**Fig. 8.** Illustration of mixed artifact normalization. Each column represents the addition of one more undersampling type on top of the previous one, and the last column shows the results after normalization by UniSino.

Such mixed artifacts pose a significant challenge to tradiational restoration strategies that adopt a one-model-for-per-artifact paradigm. A key innovation and core capability of the UniSino framework lies in its ability to effectively handle these complex scenarios. To this end, we introduce a random degradation mixing strategy during data generation, which systematically enhances both robustness and generalization. Specifically, during UniSino training, high-quality CT images and their corresponding degraded versions are paired, with each degraded sinogram generated by randomly combining artifacts from different degradation modes. As a result, the diffusion model is exposed to inputs containing one or more mixed artifacts, encouraging it to learn shared artifact characteristics and their interactions rather than simply memorizing specific patterns. Benefiting from this training strategy and from the stronger distributional consistency in the projection domain compared with the image domain, UniSino achieves reliable normalization even under multiple mixed degradations.

### D. UniSino Enables Excellent Generalization Across Multiple Cases

UniSino demonstrates excellent generalization capabilities across diverse medical imaging datasets [42-49]. Fig. 9 illustrates the overall data composition, while Table III summarizes the cross-dataset performance. The model is trained on a composite dataset that includes NLST, CQ500, and LIDC-IDRI, covering diverse anatomical regions such as head, chest and abdomen. This extensive training provides a robust foundation for handling various data characteristics.

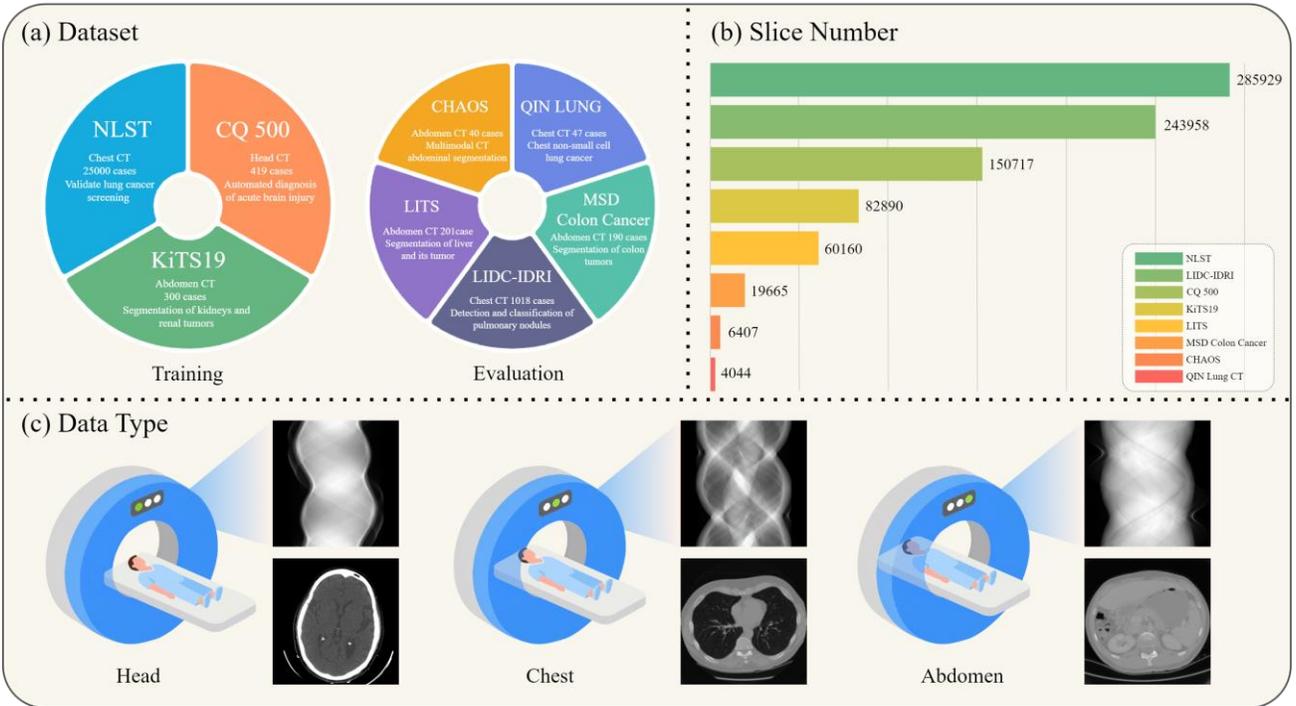

**Fig. 9.** Overview of the angular consistency loss mechanism. Given two known projection angles, their visibility ranges are computed and intersected in the image domain to infer a physically plausible projection at an unseen angle. This inferred projection is then binarized to generate a visibility mask, which is used to suppress non-physical values in the reconstructed sinogram.

**Table III**
SINOGRAM COMPARISON ON NLST DATASET

| Method | PSNR | SSIM | NRMSE |
|--------|------|------|-------|
| CycleGan | 28.71 | 79.542 | 0.08253 |
| ViT | 32.96 | 74.856 | 0.01959 |
| U-Net | 41.53 | 91.625 | 0.01254 |
| DDPM | 43.56 | 94.411 | 0.00842 |
| UniSino | 46.19 | 96.731 | 0.00331 |

To rigorously evaluate generalization, UniSino is tested on five independent datasets that are entirely unseen during training. The results show that UniSino consistently maintains high performance on these out-of-distribution datasets, highlighting its strong robustness and ability to adapt to new and varied data sources without the need for specialized retraining. Traditional models often struggle with such generalization tasks and typically require dataset-specific fine-tuning to achieve acceptable results. UniSino surpasses these conventional methods by providing a single unified framework that consistently delivers high-quality sinogram restoration across different body parts and imaging protocols. This powerful generalization underscores its potential as a foundational model for clinical applications.

### E. Ablation Studies

Traditional diffusion models, while powerful for generative tasks, often struggle with domain-specific inverse problems when lacking explicit integration of physical priors. Relying solely on data-driven patterns, they fail to fully capture the constraints governing projection-domain data, which limits reconstruction fidelity, physical plausibility, or robustness against complex artifacts.

Table IV presents the experimental results of ablation experiments on UniSino. When SinoVAE is replaced with a standard encoder-decoder structure, performance degrades, confirming that the explicit preservation of high-frequency information is critical for learning robust features. Similarly, removing SinoLoss also causes a drop in projection-domain metrics, and the lack of physical guidance leads to more artifacts after back-projection in the image domain.

**Table IV**
ABLATION COMPARISON ON NLST DATASET

| Method | (w/o) SinoVAE | (w/o) SinoLoss | UniSino |
|--------|---------------|----------------|---------|
| PSNR | 43.647 | 44. 242 | 45.522 |
| SSIM | 94.483 | 95.612 | 96.272 |
| NRMSE | 0.00791 | 0.00654 | 0.00607 |

In contrast, UniSino outperforms traditional diffusion models by embedding physical knowledge directly into its framework. SinoVAE incorporates a physics-informed priors to construct a latent space that explicitly encodes projection physics, which in turn provides a structured and physically meaningful representation for subsequent diffusion-based refinement stage. This physics constrained latent space steers the reverse diffusion process toward restorations that are not only visually plausible but also physically consistent. Quantitative evaluations further confirm that UniSino achieves superior metrics and more accurate artifact suppression across diverse scenarios, including mixed artifacts and extreme undersampling. By grounding generative modeling in projection domain physics, UniSino bridges the gap between purely data-driven learning and physics imaging constraints, setting a new benchmark for diffusion-based approaches in medical imaging where both accuracy and physical fidelity are essential.

### III. DISCUSSION

This study introduces a foundational model in the CT projection domain, which employs a perceptual compression module, SinoVAE, to extract features from undersampled sinograms. The compressed features are then optimized through the LRD module before being reconstructed by the decoder network of SinoVAE. The training strategy with mixed undersampling types endows the model with strong generalization ability, while the incorporation of SinoLoss further reinforces its standardization consistency. UniSino achieves standardization across multiple types of artifacts and demonstrates remarkable effectiveness in a wide range of applications

Unlike existing task-specific networks, our approach emphasizes cross-task adaptability, enabling robust generalization across heterogeneous acquisition conditions. The introduction of the SimSinoCT dataset further provides a benchmark resource for evaluating sinogram standardization models, offering a foundation for reproducibility and future methodological advances. Collectively, these contributions advance the development of physics-aware generative AI in CT, bridging a critical gap between handcrafted preprocessing pipelines and scalable, data-driven solutions.

By consolidating previously isolated preprocessing tasks into a single generative framework, UniSino reduces dependence on vendor-specific correction software and streamlines clinical deployment. Beyond technical integration, our experiments demonstrate that UniSino delivers consistent improvements across both single-task and mixed-task reconstruction scenarios, highlighting its generalizability under diverse acquisition conditions. This versatility suggests that a foundational model can serve not only as a universal standardizer for raw CT data but also as a transferable backbone for downstream applications, underscoring the feasibility and promise of extending foundation generative AI into the raw-data domain of medical imaging.

Despite these advances, several limitations remain. First, while UniSino demonstrates strong adaptability for CT raw-data standardization, the broader multimodal potential of diffusion-based generative models is yet to be fully explored. Extending the framework to other modalities—such as k-space representations in MRI or sinogram domains in PET—could further enhance its universality. Second, the current implementation operates primarily on two-dimensional CT sinograms, whereas clinical practice often relies on volumetric imaging. Future work will need to address the computational and memory demands of scaling UniSino to high-resolution 3D si-nogram standardization. Third, although UniSino already establishes a strong foundation in the sinogram domain by effectively suppressing error propagation, future research should further explore cross-domain and multi-domain strategies that build upon this standardized sinogram representation, thereby enabling more comprehensive artifact suppression and reconstruction consistency.

In summary, UniSino establishes a new paradigm for CT sinogram preprocessing. Its unified architecture and strong performance highlight the enormous potential of general-purpose models for raw-data standardization in medical imaging. By leveraging multi-task and multi-modal data for diverse preprocessing needs, UniSino points toward the development of AI-driven solutions tailored to specific clinical requirements. Against the backdrop of a broader shift in medical imaging from task-specific networks to universal foundation models, we envision UniSino as an effective platform capable of handling increasingly diverse and multimodal data combinations, thereby advancing the field toward more versatile and scalable imaging solutions.

## IV. METHOD

### A. Undersampled Dataset Construction

In real clinical scenarios, sinogram data is frequently affected by multiple types of undersampling that occur simultaneously. These may be caused by technical constraints, hardware limitations, or the need to reduce patient radiation exposure. To simulate such conditions in a controlled setting, we construct a comprehensive training dataset by artificially introducing a variety of undersampling patterns into standard sinograms using a self-supervised approach. The experiments obtained different types of undersampled data by applying the methods listed in Table V to the standard dataset.

**Table V**
TYPES AND MODES OF UNDERSAMPLING DATA

| Type | Equation | Description |
|------|----------|-------------|
| Low-dose | $x_{\mathrm{LD}}(s,\theta) = \ln\left[\dfrac{I_0}{Poisson(I_0\exp(-x_0))}\right]$ | low-dose noise artifact, photon starvation artifact |
| Spares-view | $x_{\mathrm{SV}}(s,\theta) = \mathrm{Mask}_{\mathrm{views}}(x_0)$ | streak artifact |
| Limited-angle | $x_{\mathrm{LA}}(s,\theta) = W(\theta_1,\theta_2) * x_0$ | slope artifact |
| Truncate | $x_{\mathrm{TR}}(s,\theta) = \begin{cases} x_0(s,\theta), s \in [s_1,s_2] \\ 0, \quad\quad \text{otherwise} \end{cases}$ | field-of-view truncation artifact |
| Metal | $x_{\mathrm{ME}}(s,\theta) = x_0 *(1\text{-}P(\mathrm{Mask}(f)))\text{+}C* P(\mathrm{Mask}(f))$ | metal-induced streak artifact, beam hardening |
| Geometry | $x_{\mathrm{GE}}(s,\theta) = x_0(s + \Delta(\theta),\theta)$ | geometric artifact, misalignment artifact |
| Ring | $x_{\mathrm{RI}}(s,\theta) = x_0(s,\theta) + a(s) * \mathrm{Mask}(s)$ | ring artifact, concentric ring pattern |
| Motion | $x_{\mathrm{MO}}(s,\theta) = P(\mathcal{H}_\theta(f))$ | patient\ respiratory\cardiac motion artifact |

Table of the formalized definitions of undersampling types. $x$ denotes the sinogram, $s$ represents the number of sensors, and $\theta$ indicates the scanning angle.

Specifically, we simulate sparse view sampling by removing a subset of projection angles, which mimics faster scan protocols. Limited-angle acquisition is modeled by restricting the angular range of projections, replicating the constraints of interventional procedures or hardware limitations. Truncation artifacts are introduced by cropping the sinogram along the detector axis to emulate scenarios where the scanned object exceeds the field of view. We simulate detector downsampling by subsampling the detector elements, which reflects reduced detector resolution or malfunction. Low-dose conditions are generated by injecting Poisson noise into the sinogram, imitating reduced exposure settings used to limit radiation. We introduce ring artifacts by assigning consistent errors to certain detector channels, which simulates calibration failures. Geometric distortions are applied by perturbing system parameters such as the rotation center or source-detector alignment. Finally, to replicate metal artifacts, we simulate high-attenuation regions by zeroing out or distorting projection values that intersect metallic implants.

These simulated artifacts are randomly combined at varying ratios, resulting in a diverse dataset that captures the complex and heterogeneous nature of real-world undersampled data. By training on such a dataset, UniSino is exposed to a broad spectrum of corruption types, which facilitates its ability to generalize and adapt to unseen conditions. The use of synthetic but physically meaningful artifacts also allows for scalable data generation without manual labeling, which makes this approach suitable for large-scale pretraining.

### B. SinoVAE with High Frequency Extraction

To enable robust sinogram standardization across diverse inputs, UniSino adopts an encoder-decoder framework with high frequency extraction. The architecture integrates a novel variational autoencoder that compresses sinogram data into a structured latent space. The encoder is designed as a variational autoencoder with KL divergence regularization. It extracts latent features from input sinograms into a smooth and continuous embedding space. The encoder $E$ maps the input high-quality sinogram $x$ to a latent representation $z$. This process can be expressed as:

$$(\mu, \sigma) = E(x), \qquad (1)$$

In this latent space, each sample is represented by a distribution that captures the underlying content of the input sinogram. The latent space is regularized by KL divergence, ensuring it approximates a standard normal distribution. The latent variable $z$ is sampled using the reparameterization trick:

$$z = \mu + \sigma \odot \epsilon, \text{ where } \epsilon \sim \mathcal{N}(0, I), \qquad (2)$$

where $\mu$ and $\sigma$ are the mean and standard deviation vectors produced by the encoder. The training of SinoVAE is designed to maximize the similarity between the input sinogram and its reconstruction. This is achieved by maximizing the evidence lower bound, which serves as a tractable objective for the data log-likelihood. The training process encourages the decoders to produce an output that faithfully matches the original input $x$.

A key innovation is the dual-decoder structure. This structure includes a global decoder and a high-frequency decoder to reconstruct the sinogram from the latent embedding. The global decoder is responsible for reconstructing the overall structure of the sinogram. Its design incorporates a physically informed module that utilizes both a sinogram loss and a reconstruction loss to ensure physical consistency. The high-frequency decoder focuses on fine details and edges. This decoder is guided by a conditional edge loss. To extract the high-frequency information required for this loss, a Sobel operator is applied to the original sinogram. The Sobel operator computes the image gradient magnitude $G$ from the horizontal and vertical gradients $G_x$ and $G_y$:

$$G_x = \begin{bmatrix} 1 & 0 & -1 \\ 2 & 0 & -2 \\ 1 & 0 & -1 \end{bmatrix} * x, \; G_y = \begin{bmatrix} 1 & 2 & 1 \\ 0 & 0 & 0 \\ -1 & -2 & -1 \end{bmatrix} * y, \qquad (3)$$

$$G = \sqrt{G_x^2 + G_y^2}, \qquad (4)$$

where $*$ denotes the 2D convolution operation.

Furthermore, both decoders are augmented with a discriminator network. The discriminators help to enforce realism and consistency in the outputs of both the global and high-frequency pathways. This entire framework allows the model to learn a comprehensive representation that separates global anatomical structures from fine-grained textural details, leading to superior sinogram standardization and reconstruction. To ensure that the generated sinograms are physically consistent and visually faithful, the training process also incorporates a structural consistency loss and an artifact suppression loss. The structural consistency loss penalizes discrepancies in gradient features, which promotes accurate edge preservation. The artifact suppression loss discourages the network from retaining characteristic patterns of noise and corruption.

### C. Latent Refinement Diffusion

Diffusion models represent a class of generative models that have recently demonstrated state-of-the-art performance in high-fidelity image synthesis. Our framework leverages a diffusion mechanism in its second stage to refine the coarse latent representations generated by SinoVAE. This stage is critical for restoring high-frequency details and mitigating artifacts introduced by undersampling, thereby enabling high-quality sinogram reconstructions from severely degraded measurements. The core principle is to reverse a Markov chain that gradually adds noise to a clean latent vector, effectively learning to denoise a random Gaussian vector into a structured, refined latent representation.

The diffusion process consists of two opposing procedures: a fixed forward noising process and a learned reverse denoising process. The forward process, denoted by $q$, systematically corrupts an initial clean latent vector $z_0$, by adding Gaussian noise over $T$ discrete time steps. This is defined as a Markov chain:

$$q(z_t | z_{t-1}) = N(z_t; \sqrt{(1 - \beta_t)} z_{t-1}, \beta_t I), \qquad (5)$$

where $\beta_t$ is a small positive constant representing the noise schedule at step $t$. A key property of this process is that we can sample $z_t$ at any arbitrary timestep $t$ directly from $x_0$ in a closed form. Letting $\alpha_t = 1 - \beta_t$ and $\bar{\alpha}_t = \prod_{i=1}^{t} \alpha_i$, the distribution of $z_t$ given $z_0$ is:

$$q(z_t | z_0) = N(z_t; \sqrt{\bar{\alpha}_t} z_0, (1 - \bar{\alpha}_t)I). \qquad (6)$$

This allows us to express any noisy latent $z_t$ as $z_t = \sqrt{\bar{\alpha}_t} z_0 + \sqrt{(1 - \bar{\alpha}_t)} \varepsilon$, where $\varepsilon$ is a standard Gaussian noise vector. The reverse process $p_\theta$, aims to reverse this corruption, starting from a pure noise vector $z_t \sim N(0, I)$ and iteratively denoising it to produce a refined latent vector $x_0$. This process is also modeled as a Markov chain, conditioned on the latent representation $c$ of the undersampled sinogram, which is provided by the frozen SinoVAE encoder. The transitions of this reverse chain are parameterized by a neural network, $\varepsilon_\phi(z_t, t, c)$, which is trained to predict the noise component $\varepsilon$ from the noisy latent $x_t$ at time $t$ and condition $c$. The iterative reconstruction of latent variable at each step is formulated as:

$$z_{t-1} = \frac{1}{\sqrt{\alpha_t}}(z_t - \frac{1 - \alpha_t}{\sqrt{(1 - \bar{\alpha}_t)}} \varepsilon_\phi(z_t, t, c)) + \sigma_t \varepsilon, \qquad (7)$$

where $\varepsilon$ is a standard Gaussian noise vector and $\sigma_t$ is a scaling factor for the noise. This equation forms the fundamental building block of the generative process. Starting with $z_t$, the model predicts the noise that was added to obtain $z_t$ and subtracts it to estimate a less noisy $z_{t-1}$. This is repeated $T$ times to yield the final refined latent $z_0$.

The neural network $\varepsilon_\phi$ is trained using a simplified objective derived from the variational lower bound on the data log-likelihood. This objective simplifies to a mean squared error loss between the true and predicted noise. By minimizing this objective, the model learns to effectively capture the conditional distribution of high-frequency details given the coarse global structure. Upon completion of the $T$ reverse steps, the resulting refined latent $x_0$ is fed into the frozen SinoVAE decoder, which synthesizes the final high-resolution, artifact-free sinogram.

### D. Optimization algorithm

The training of UniSino is divided into two distinct stages to effectively disentangle latent representation learning from the generative modeling of sinograms.

In the first stage, the encoder-decoder backbone is trained as a SinoVAE, in an unsupervised fashion. The training objective is not only to reconstruct the original sinogram from its latent representation but also to ensure that the latent space is regularized and perceptually meaningful. To achieve this, we employ a comprehensive loss function that combines multiple objectives.

Let $S$ denote the ground-truth sinogram. The encoder $E$ maps the sinogram into a latent representation $z = E(S)$, while the decoder $D$ attempts to reconstruct the sinogram from $z$. The VAE objective in this phase is a weighted sum of a reconstruction loss, a perceptual loss, a KL-divergence term for latent space regularization, and an adversarial loss:

$$L_{SinoVAE} = L_{rec}(D(z), S) + \lambda_p L_{per}(D(z), S) + \beta L_{KL}\big(q(z|S)||p(z)\big) + \lambda_{adv} L_{adv}\big(D(z)\big), \quad (8)$$

where $L_{rec}$ is the L2 reconstruction loss enforces pixel-level fidelity. The perceptual loss $L_{per}$ calculated using the LPIPS metric, ensures that the reconstructions are perceptually similar to the ground truth. The KL-divergence term $L_{KL}$ regularizes the output of encoder to follow a standard normal distribution, creating a smooth and continuous latent space. Finally, the adversarial loss $L_{adv}$, driven by a patch-based discriminator, pushes the model to generate reconstructions that are indistinguishable from real sinograms, enhancing their realism and sharpness.

In the second stage, the diffusion model is trained in the latent space using the encoder output from corrupted sinograms as conditional guidance. Specifically, we construct corrupted sinograms $S_{corr}$ from clean sinograms by applying synthetic undersampling transformations, and then encode them to $z_{cond} = E(S_{corr})$. This conditional denoising process is trained using the standard DDPM loss adapted to the latent space. The overall optimization objective is:

$$L_{diffusion} = \mathbb{E}_{z_t, t}\left[\left\| \varepsilon_\phi(z_t, t, z_{cond}) - \varepsilon \right\|_2^2\right], \quad (9)$$

where $z_t$ is the noisy latent variable at timestep $t$, $\epsilon$ is the true noise, and $\varepsilon_\phi$ is the noise predicted by the diffusion model. This two-stage strategy allows UniSino to first build a robust and physically-informed representation of sinograms, and then leverage it during generative diffusion to perform high-fidelity restoration under various types of undersampling.